\documentclass{article}
\usepackage{cite}
\usepackage{a4wide}
\usepackage{amsfonts}
\usepackage{amsmath}
\usepackage{algorithm}
\usepackage{algorithmic}
\usepackage{caption}
\usepackage{booktabs}  
\usepackage{graphicx}
\usepackage[caption=false,font=footnotesize]{subfig}
\usepackage{url}
\usepackage[colorinlistoftodos]{todonotes}

\begin{document}

\title{EvoMSA: A Multilingual Evolutionary Approach for Sentiment Analysis}
\author{Mario Graff$^{1,3}$ \and Sabino Miranda-Jiménez$^{1,3}$ \and Eric S. Tellez$^{1,3}$ \and Daniela Moctezuma$^{1,2}$}
\date{%
$^1$INFOTEC Centro de Investigaci\'on e Innovaci\'on en Tecnolog\'ias
de la Informaci\'on y Comunicaci\'on, Circuito Tecnopolo Sur No 112, Fracc. Tecnopolo Pocitos II, Aguascalientes 20313, M\'exico\\
$^2$CentroGEO Centro de Investigaci\'on en Ciencias de Informaci\'on Geoespacial,
Circuito Tecnopolo Norte No. 117, Col. Tecnopolo Pocitos II, C.P., Aguascalientes, Ags 20313 M\'exico\\
$^3$CONACyT Consejo Nacional de Ciencia y Tecnolog\'ia,
Direcci\'on de Cátedras, Insurgentes Sur 1582, Cr\'edito Constructor, Ciudad de M\'exico 03940 M\'exico~\\ ~\\
This work has been accepted for publication to the~\\
IEEE Computational Intelligence Magazine~\\
\copyright 2019 IEEE.  Personal use of this material is permitted.  Permission from IEEE must be obtained for all other uses, in any current or future media, including reprinting/republishing this material for advertising or promotional purposes, creating new collective works, for resale or redistribution to servers or lists, or reuse of any copyrighted component of this work in other works.
}

\maketitle

\begin{abstract}
Sentiment analysis (SA) is a task related to understanding people's feelings in written text; the starting point would be to identify the polarity level (positive, neutral or negative) of a given text, moving on to identify emotions or whether a text is humorous or not. This task has been the subject of several research competitions in a number of languages, e.g., English, Spanish, and Arabic, among others. In this contribution, we propose an SA system, namely EvoMSA, that unifies our participating systems in various SA competitions, making it domain-independent and multilingual by processing text using only language-independent techniques. EvoMSA is a classifier, based on Genetic Programming that works by combining the output of different text classifiers to produce the final prediction. We analyzed EvoMSA on different SA competitions to provide a global overview of its performance. The results indicated that EvoMSA is competitive obtaining top rankings in several SA competitions. Furthermore, we performed an analysis of EvoMSA's components to measure their contribution to the performance; the aim was to facilitate a practitioner or newcomer to implement a competitive SA classifier. Finally, it is worth to mention that EvoMSA is available as open-source software.
\end{abstract}

\maketitle

\section{Introduction}

Sentiment Analysis (SA) is a task dedicated to developing automatic techniques that can analyze people's feelings or beliefs expressed in texts~\cite{Liu2012AAnalysis} such as emotions, opinions, attitudes, appraisals, among others. Sentiment Analysis is not only applied to text data but also voice, video recording, to mention a few, see, for instance \cite{Poria2016FusingContent} and \cite{Ortis2018VisualImages}. Regarding text, one of the most analyzed opinion forums is Twitter because it is a massive source of data\footnote{https://www.omnicoreagency.com/twitter-statistics/} having potential uses for many decision-making areas. Affective computing and sentiment analysis have attracted a multitude of researchers aiming to understand people's opinion on an event or entity or even the user's mood~\cite{Liu2012AAnalysis,Cambria2016}. The dedicated community, i.e., researchers in areas ranging from psychology and sociology to natural language processing (NLP) and machine learning, have proposed a number of approaches to tackle the problem. The community also organizes several challenges to measure the effectiveness of the available approaches over common ground, e.g., TASS (Taller de An\'alisis Sem\'antico)~\cite{Martinez-Camara2018OverviewEmotions,Martinez-Camara2017Overview2017} and SemEval (Semantic Evaluation)~\cite{Mohammad2018SemEval-2018Tweets,Rosenthal2017SemEval-2017Twitter} which are among the most popular SA competitions. Overall, the challenges' dynamic provides insightful ideas on how to solve the problem, and an objective procedure to compare different approaches; however, in our opinion, the side effect is that some models are difficult to replicate.  Our particular experience is that the rush of the competition leads us to take several decisions which are not systematically tested. Consequently, it produces many details that are impractical to write in a report. 

One would expect that in competitions such as SemEval~\cite{Mohammad2018SemEval-2018Tweets,Rosenthal2017SemEval-2017Twitter} where a task is in English, Arabic, and Spanish languages there would be plenty of multilingual approaches participating in all the languages, or, at least, teams participating in various languages. However, the majority of systems are designed to work only in English. For example, in SemEval 2017~\cite{Rosenthal2017SemEval-2017Twitter} where task 4 was in English and Arabic only 19\% (8 out of 42 teams) of the teams participated in both languages, and in SemEval 2018~\cite{Mohammad2018SemEval-2018Tweets} 16\% (7 out of 43 teams) participated in English, Arabic, and Spanish, and 28\% (12 out of 43) participated in two of the languages. The reason for the reduced multilingual participation is the inherent difficulties of creating multilingual systems. For example, the resources are only available in a specific language, or the implementation of other languages is challenging. This problem becomes relevant for those languages with weakly developed NLP techniques. As an example, some winning approaches have created text models based on millions of texts (more than 400 million of tweets), clearly, the requirements on information and computing power limit this approach only to the languages where these requirements are satisfied which sometimes are those where the authors have invested most of their time. On the other hand, there are a number of language techniques that are tailored to a specific language, and, to target another language, one needs to be fluent on that particular language.

To overcome these problems, this contribution proposes a multilingual methodology that tackles the sentiment analysis task inspired by our participation as INGEOTEC in TASS 2017~\cite{Moctezuma2017ATASS17} and 2018~\cite{Moctezuma2018INGEOTECCompetition}; SemEval 2017~\cite{Miranda-Jimenez2017INGEOTECAnalysis} and 2018~\cite{Graff2018}; and IberEval 2018 in MEX-A3T~\cite{Graff2018a}, and HAHA~\cite{Ortiz-Bejar2018} competitions. There are considerable differences between INGEOTEC's systems and EvoMSA. Firstly, EvoMSA is applied to all the languages and competitions without any modification and with its parameters fixed, per language, to provide a global overview of its performance; whereas, INGEOTEC systems are slightly different in each competition. Secondly, in this contribution, it is included an alternative implementation of the DeepMoji~\cite{Felbo2017UsingSarcasm}, ad-hoc to our approach; we call it Emoji Space. Finally, some text models have not been used in our participating systems such as FastText (except in TASS 2018) and our Emoji Space (see Section~\ref{sec:description}).

The goal is to propose a competitive multilingual SA system that can be applied to a variety of languages and domains. To achieve this, we disregard those techniques and optimizations that are either only applicable to a particular language or domain, or which net effects (regarding performance) are hard to measure. Moreover, the development of EvoMSA is modular so that each of its parts can be measured separately, facilitating the understanding of which parts contribute the most to the performance. As a result, the methodology presented here can be easily applied to other text categorization problems, and it is easy to implement, given that there are public libraries for most of its components. Moreover, we released our Python implementation as open-source\footnote{\url{https://github.com/INGEOTEC/EvoMSA}}.

The rest of the manuscript is organized as follows: Section~\ref{sec:related-work} presents the related work emphasizing the best or multilingual works presented at SA competitions. EvoMSA is described in Section~\ref{sec:description}. Section~\ref{sec:datasets} describes the competitions datasets used as testbeds. The performance and comparison of EvoMSA using different models and state-of-the-art SA systems are described in Section~\ref{sec:results}. The conclusions and possible directions for future work are given in Section~\ref{sec:conclusions}.

\section{Related Work}
\label{sec:related-work}
 
The sentiment analysis community has stimulated research groups to develop innovative techniques to classify aspects, stances, emotions employing international challenges such as SemEval, TASS, IberEval, among others. In particular, to boost multilingual approaches, SemEval challenge encourages the participation in more than one language; for instance, English, Spanish, and Arabic languages are promoted in tasks such as polarity detection~\cite{Rosenthal2017SemEval-2017Twitter} and emotion detection~\cite{Mohammad2018SemEval-2018Tweets}.

%%%%%%%%%%% Multilingual approaches

Existing multilingual approaches rely on lexicons, parallel corpora, machine translation systems, labeled data, or a combination of them~\cite{Lo2017499,vilares2017}. For example, polarity detection \cite{Balahur2014} uses a machine translation system to translate data from English into four languages (Italian, German, French, and Spanish), and a classifier to train models for each language. The results by language are similar, and the combination of multilingual data sometimes improves the performance; the authors also point out that the use of external labeled data of the target language improves the performance. Meng {\it et al}. ~\cite{Meng:2012} proposed a generative cross-lingual mixture model (CLMM) using the bilingual parallel corpus for English and Chinese (target), they can learn unseen sentiment words maximizing the likelihood of generating the parallel corpora. Becker {\it et al}. \cite{becker2017} used two source corpora (English and Portuguese) of news and their translated versions of the target languages Spanish, French, English, and Portuguese. The combination of features of multilingual translations improves the performance for the classification task; on the other hand, the stacking of monolingual classifiers performs even better. 

%%%%%%%%%%%%%%%%%%%%%%%%%%%%%%% SemEval 2017

In the case of contests, we describe those approaches that obtained the first positions in each competition, in SemEval 2017, polarity detection task involves detecting whether a given text has a positive, negative, or neutral sentiment at a global level. The BB\_twtr~\cite{ClicheBloomberg2017BBLSTMs} team used an ensemble of Neural Networks combining Convolutional Neural Networks (CNNs) and Long-Short Term Memory Networks (LSTMs). The DataStories~\cite{Baziotis2017DataStoriesAnalysis} system follows a similar deep learning approach using Bidirectional LSTMs (BiLSTM) with an attention mechanism. Both approaches use word embeddings from pre-trained vectors as text representation. In the case of the Arabic language, NileTMRG team~\cite{El-Beltagy2017NileTMRGAnalysis} used a Naive Bayes classifier augmented with phrase and word level sentiment lexicon for Egyptian and Modern Standard Arabic. Two multilingual systems were proposed for this task, SiTAKA~\cite{Jabreel2017SiTAKAFeatures} and ELiRF-UPV~\cite{Gonzalez2017ELiRF-UPVLearning} which participated in English and Arabic. SiTAKA system uses pre-trained embeddings, Word2Vec for English, and SKIP-G300~\cite{Zahran2015WordArabic} for Arabic. This system also uses other features such as $n$-words, part of speech tags, and lexicons to give an additional score. It uses a Support Vector Machine (SVM) to perform the classification. ELiRF-UPV system is based on Convolutional Recurrent Neural Networks (CRNNs) and the combination of general and specific word embeddings for English and Arabic, and polarity information from lexicons.

%%%%%%%%%%%%%%%%%%%%%%%%%%%%%%% SemEval 2018

SemEval 2018~\cite{Mohammad2018SemEval-2018Tweets} consisted of an array of subtasks where the systems have to infer the emotional state of a person based on his/her tweets. The tasks include the automatic determination of emotion intensity (EI) and valence classification (VC). The former tries to determine the emotional intensity of tweets; it considers four basic emotions: anger, fear, joy, and sadness. The latter, VC, consists on, given a tweet, classify it into one of seven ordinal classes related to various levels of positive and negative sentiment intensity. All tasks were run for English, Arabic, and Spanish languages. In this competition, SeerNet system~\cite{Duppada2018SeerNetTweets}, participating only in English, proposed a pipeline of pre-processing and feature extraction steps. The pre-processing uses Tweettokenize\footnote{https://github.com/jaredks/tweetokenize} tool, and for feature extraction, several deep learning approaches were considered, such as DeepMoji, EmoInt, Sentiment Neuron, and Skip-Thought Vectors. EiTAKA~\cite{Jabreel2018EiTAKATweets} presented results for English and Arabic using an ensemble of two approaches, deep learning and XGBoost regressor based on embeddings and lexicons. As a multilingual system, AffecThor~\cite{Abdou2018AffecThorTweets} participated in all the languages and emotional intensity and valence task. The AffecThor team proposed a solution build upon several best past-years participating systems and a combination of several approaches based on lexical resources and semantic representations. These resources include 22 lexicons and Word2Vec for word embeddings. In the classification step, they use the architecture of several neural models like CNN with max pooling, BiLSTM with attention, and a set of character and word features BiLSTMs (CHAR-LSTM). 

%%%%%%%%%%%%%%%%%%%%%%%%%%%%%%% TASS 2017

TASS 2017 competition \cite{Martinez-Camara2017Overview2017} focused on polarity classification at tweet level (positive, negative, neutral, and none) in the Spanish language. The systems were evaluated on two datasets: the International TASS corpus (InterTASS), tweets located inside Spain territory written in the Spanish language; and the General Corpus, tweets of personalities and celebrities written in Spanish from several countries including Spain. ELiRF-UPV~\cite{Hurtado2017ELiRF-UPVLearning} employed different approaches, i.e.,  bag-of-words, bag-of-chars, word embeddings, and one-shot vectors over words and characters representations, as well as, Multilayer Perceptron (MPL), RNNs, CNNs, and LSTM networks.

%%%%%%%%%%%%%%%%%%%%%%%%%%%%%%% TASS 2018

TASS 2018 edition~\cite{Martinez-Camara2018OverviewEmotions} proposed tasks including the identification of positive or negative emotions that can arouse in news, i.e., classify news articles into {\em SAFE} (positive emotions, so safe for ads) or {\em UNSAFE} (negative emotions, so better avoid ads), as a kind of stance classification according to reader's point of view. In this task, there were two subtasks, subtask 1 (S1) consists in the classification of headlines into either SAFE or UNSAFE tweets written only in the Spanish language spoken in Spain; there were two test sets, named L1 and L2, having as only difference their cardinality. Moreover, subtask 2 (S2) consists in evaluating the systems' ability to generalize. For training, participants were provided with headlines written only in the Spanish language spoken in Spain, and for testing, news articles come from nine different countries of America in order to encourage generalization. ELiRF-UPV~\cite{Gonzalez2018:ELiRF-UPV} team used a deep neural network, Deep Averaging Networks (DANs), and a set of pre-trained word embeddings for representing the news headlines.

%%%%%%%%%%%%%%%%%%%%%%%%%%%%%%% MEX-A3T

The IberEval contest is related to emotions, mostly in the Spanish language. In its 2018 edition, IberEval promoted different tasks such as aggressiveness identification~\cite{Alvarez-Carmona2018OverviewTweets} and humor analysis~\cite{Castro2018Overview2018}. The aggressiveness identification task (MEX-A3T) is motivated by cyberbullying, hate speech, harassment, among others. It consists of classifying a text, in Spanish from Mexico, into either aggressive or non-aggressive. CGP~\cite{EnriqueMunizCuza2018AttentionDetection} system used an Attention-based LSTM network, and word embeddings were used over the sentence. Attention is applied over the hidden states to estimate the importance of each word, and this context vector is used into another LSTM model to estimate whether a tweet is aggressive or not. Aragon-Lopez~\cite{Aragon2018Author2018} team used both a bag of terms representation and second-order attributes (SOA). They use an $n$-gram representation combined with a CNN as the classifier.

%%%%%%%%%%%%%%%%%%%%%%%%%%%%%%% HAHA 2018

The HAHA task \cite{Castro2018Overview2018} (Humor Analysis based on Human Annotation) consisted of classifying tweets in Spanish as humorous or not.  U\_O-UPV~\cite{Ortega-Bueno2018UOMedia} used a neural network with attention mechanism, word2vec models, and a set of linguistic features such as stylistic (e.g., length, counting of emoticons, hashtags), structural and content (e.g., animal vocabulary, sexual and obscene vocabulary), and affective (e.g., positive or negative words, counting of words related to attitudes). 
%%%%%%%%%%%%%%% Sarcasm
The use of different Neural Networks has not been restricted to the aforementioned tasks. There are essential advances on tasks such as in Sarcasm which could be considered as a verbal form of irony that toggles the explicit sentiment found in a text \cite{Joshi2017AutomaticDetection}. Joshi {\it et al}.~\cite{Joshi2016AreDetection} proposed a deep-learning approach with word embeddings as the main feature. Another deep-learning approach was presented in \cite{Amir2016ModellingMedia}; here, authors use CNN to learn user embeddings with the purpose to learn user-specific context. Ghosh and Veale \cite{Ghosh2016FrackingNetwork} explored and compared the performance of CNN and RNN regarding sarcasm detection.

%%%%%%%%%%%%%%%%%%%%%%%%%%%%%%%%%%%%%%%%%%%%%%%%%%%%%%%%%%%%%%%%%%%%%%%%%%%%%%%%%%%%%%%%%%%%%%%

\section{System Description}
\label{sec:description}
\begin{figure}[ht!]
	\centering\includegraphics[width=0.85\textwidth]{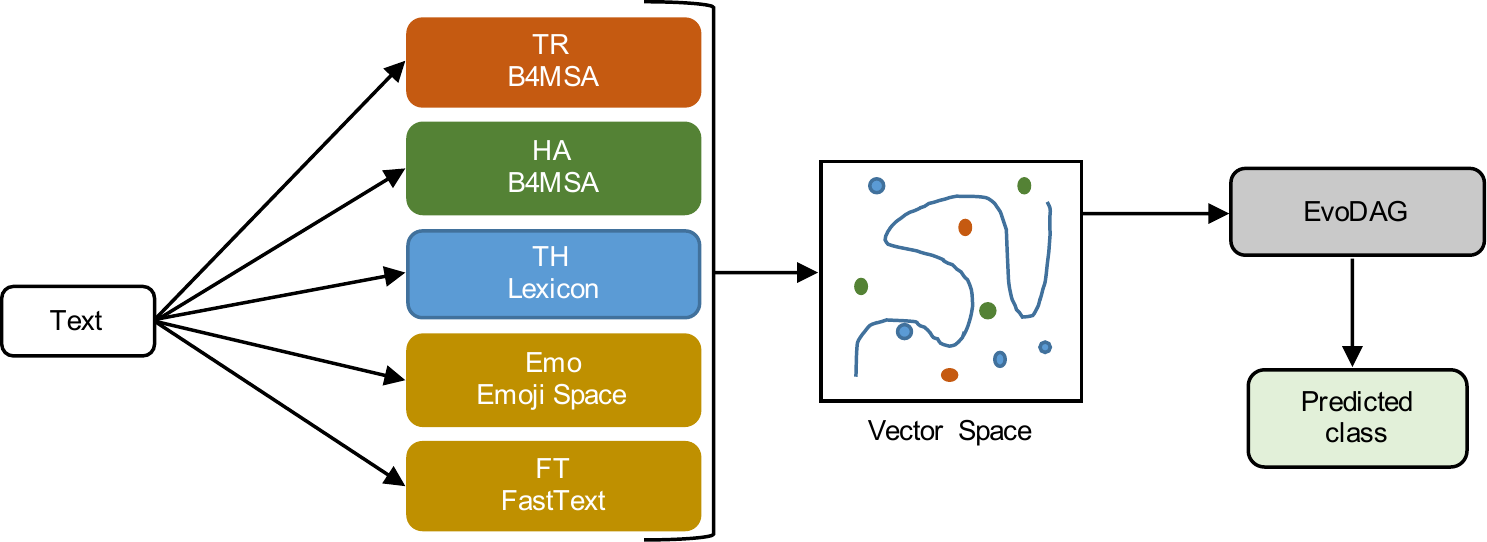}
	\captionsetup{justification=centering}
	\caption{The prediction scheme of EvoMSA. A text is transformed by different models composing a vector space which is the input of EvoDAG to make the final prediction; the flow goes from left to right.}
	\label{fig:evodag}
\end{figure}

EvoMSA is a specialization of Stack Generalization (SG) \cite{Wolpert1992StackedGeneralization} focused on text classification problems\footnote{Stack Generalization was initially proposed to improve the performance of any supervised learning algorithm.}. EvoMSA is a two-stage procedure where the first stage is composed by several models that transform a text into decision function values; these values are combined, in the second stage, by a classifier, in particular, EvoDAG~\cite{Graff2016,Graff2017} which is based on Genetic Programming (GP). Figure~\ref{fig:evodag} depicts the structure of EvoMSA; the prediction flow goes from left to right. On the left, a text is submitted to different models; the outputs of these models compose the vector space which is used by EvoDAG to make the final prediction. Figure~\ref{fig:evodag} illustrates that the difference between EvoMSA and SG is on the first stage, whereas EvoMSA's first stage receives a text, SG receives a vector; from the vector space through the end of the procedure EvoMSA and SG are equivalent. 

The first stage considers five models, that can be selected by the user, and are a composition of two functions, i.e., $g \circ m$, where $m$ transforms a text into a vector (i.e., $m: \text{text} \rightarrow \mathbb{R}^d$ where $d$ is inherent to $m$) and $g$ is a function with the form $g: \mathbb{R}^d \rightarrow \mathbb{R}^c$ where $c$ is the number of classes of the text classification problem. Function $m$ is a text model obtained from different sources, and $g$ is a linear SVM\footnote{It was decided to use a linear SVM as $g$ based on our experience building B4MSA~\cite{Tellez2017ATwitter}, our previous SA classifier.}.

The different sources used to compute $m$ are: in the first model, $m_1$, the training set of the competition (TR). The second model, $m_2$, uses a human annotated (HA) dataset, independent of TR. The third model, $m_3$, is an emotion and sentiment Lexicon-based model (TH). The fourth model, $m_4$, is our Emoji Space (Emo). Finally, the fifth model, $m_5$, corresponds to FastText (FT), frequently used to provide a semantic representation (word embeddings) of the text. Based on this description, it is possible to infer the value of $d$ for each $m$. In the first model, $d_1$ corresponds to the vocabulary size. In the second model $d_2$ is the number of classes of HA. The third model $d_3$ equals $2$ which corresponds to the count of positive and negative words, respectively. The fourth model has a $d_4=64$ (see Section~\ref{sec:EmojiSpace}); and finally, $d_5=300$ (see Section~\ref{sec:FastText}) for the fifth model. 

The second stage starts using the outputs generated by all the models, e.g., $g \circ m_1, g \circ m_2, \ldots, g \circ m_\ell$, concatenated to form a vector in $\mathbb{R}^{\ell c}$ where $\ell$ is the number of models, and $c$ is the number of classes. This last vector is used by EvoDAG to perform the final prediction. Before EvoDAG can be used, it requires to be trained. The naive approach would be to use TR to train $g$ and EvoDAG. Nonetheless, this would result in an ill-designed approach, which is not considering the weakness of the classifier(s) at generalization. In SG, it was proposed to train the second stage classifier (e.g., EvoDAG) by using the output of a $k$-fold (five folds) cross-validation approach on TR and $g$. Algorithm \ref{alg:train-evodag} presents the procedure used to train EvoDAG. It receives the first-stage text models, $m \in \mathcal M$, and TR. From lines 2-9, it iterates for the different text models, $m$, transforming the text into vectors (line 3), these vectors are used in $k$-fold cross-validation (lines 5-8) to predict the decision function values of the validation set ($vs$). During the folding process, there are two disjoint sets, $tr$ and $vs$, where $tr$ is used to train an SVM (line 6), and $vs$ is the set to be predicted (line 7). The predictions obtained for the different models, $\mathcal M$, are concatenated (line 9) to form EvoDAG's training set. The last step is to train EvoDAG (line 11) with the predicted values. 

\begin{algorithm}
\caption{EvoDAG's training}
\begin{algorithmic}[1]
\REQUIRE $\mathcal M$ \COMMENT{First-stage text models}
\REQUIRE TR \COMMENT{Training set of pairs text and class}
\STATE $\mathcal X \leftarrow []$ \COMMENT{Store the decision function values}
\FORALL{$m \in \mathcal M$}
    \STATE $X \leftarrow m($TR$)$ \COMMENT{Transform the texts into vectors}
    \STATE $\hat X \leftarrow []$ \COMMENT{List containing $m$ decision function values}
    \FOR{$(tr, vs) \in \textsf{K-Fold}(X)$}
        \STATE $c \leftarrow \textsf{Train}($SVM$, tr)$ \COMMENT{Train a linear SVM}
        \STATE $\hat X[\textsf{Index}(vs)] \leftarrow \textsf{Decision-Function}(c, vs)$
    \ENDFOR
    \STATE $\mathcal X \leftarrow \textsf{Concatenate}(\mathcal X, \hat X)$
\ENDFOR
\RETURN $\textsf{Train}($EvoDAG$, \mathcal X)$
\end{algorithmic}
\label{alg:train-evodag}
\end{algorithm}

The rest of this section describes the different text models, $m$, used in this contribution. It starts with B4MSA using two datasets, the lexicon-based models, Emoji Space and FastText. The last subsection is devoted to describing EvoDAG, the classifier used in EvoMSA's second stage.

\subsection{B4MSA}
The first two text models, i.e., $m_1$ and $m_2$, use our baseline for multilingual sentiment analysis, namely B4MSA\footnote{https://github.com/INGEOTEC/b4msa}~\cite{Tellez2017ATwitter}. B4MSA uses an equivalent structure that the models used in EvoMSA's first stage, i.e., $g_b \circ m_b$. Function $m_b$ uses a series of simple language-independent text transformations to convert text into tokens, as well as some language-dependent transformation commonly implemented on various open-source libraries. Nonetheless, it avoids the usage of computational expensive linguistic tasks such as part-of-speech tagging, dependency parsing, among others. Then, these tokens are represented into a vector space model using TF-IDF, and, finally, the vectors and their associated classes are learned by a linear SVM (i.e., $g_b$). 

B4MSA was conceived to serve as a baseline for text categorization. To achieve this, it starts with a search in its parameter space to find an acceptable configuration. However, this search, per problem, increment the time required to find a model, and besides, our previous work on sentiment analysis (see~\cite{Tellez2017AAnalysis}) indicates that some parameters could be fixed with a minimal impact on the performance. Consequently, it was decided to keep constant the parameters of B4MSA per language. 

Table~\ref{tab:b4msa-parameters} shows B4MSA's parameters per language. These parameters were obtained by measuring their performance (using macro-F1) on all the datasets used in this contribution, and, using $k$-fold cross-validation ($k=5$) on the training set. The parameter space was sampled using a loop of two steps. In the first step, the parameters varied were the tokenizers; it was tested all the combinations of $n$-words $1, 2$, and $3$; skip-grams $(3, 1), (2, 2)$, and $(2, 1)$; and $q$-grams $2, 3, 4, 5$, and $6$. The second step tested the rest of the parameters shown in the table; these parameters are either dichotomic or treated as such, this is the case of parameters with possible values like {\em group} or {\em delete}. This process continues until a stable configuration is found, that is, where the best configuration is the one found in the previous step. 

Some of B4MSA's parameters are self-described such as remove diacritics, duplicates, punctuation symbols, and convert text to lowercase. The emoticons were changed to the words {\em \_pos}, {\em \_neg}, or {\em \_neu} depending on the polarity expressed. Numbers, URLs and users are either deleted or replaced with words {\em \_num}, {\em \_url}, and {\em \_usr}, respectively. The tokens are words, bi-grams of words, $q$-grams of different sizes, and skip-grams. The notation used in skip-gram is $(a, b)$ where $a$ indicates the number of words and $b$ indices the length of the skip, for example, in {\em have a nice weekend} the skip-gram $(2, 1)$ would be {\em have nice} and {\em a weekend}.

\begin{table}[!h]
\centering
\caption{B4MSA parameters used per language.}
	\begin{tabular}{l l l l l}
	\toprule
	\multicolumn{5}{c}{\bf{Text transformation}} \\
	Parameter & Default & Arabic & English & Spanish \\
	\midrule
			remove diacritics  & yes & yes & no & yes\\
		remove duplicates  & &  & yes &\\
		remove punctuation & &  & yes &\\
		lowercase    & &  & yes &\\		
		emoticons    & &  & group &\\
		numbers    & group & group & delete & group \\
		urls       & &  & group &\\
		users      & &  & group &\\
		hashtag   & &  & none &\\
		entities   & none & delete & none & none\\
		negation   & &  & false &\\
		stopwords  & false & delete & false & false \\
		stemming   & &  & false &\\
	\midrule
	\multicolumn{5}{c}{\bf{Tokenizers}} \\
	\midrule
		$n$-words    & $\{1, 2\}$ & $\{1\}$ & $\{1, 2\}$ & $\{1\}$\\
		skip-grams & $\{\}$ & $\{\}$ & $\{(3, 1)\}$ & $\{(2, 1)\}$\\
		$q$-grams    & $\{2, 3, 4\}$ & $\{2, 3, 4\}$ & $\{3, 4\}$ & $\{2, 3, 4, 5, 6\}$\\
	\bottomrule
	\end{tabular}
\label{tab:b4msa-parameters}
\end{table}

B4MSA is used to create two models ($g \circ m_1$ and $g \circ m_2$), one using the competition training set (TR) and the other using a human annotated (HA) dataset. Regarding TR, $m_1=m_b$, i.e., $m_1$ is B4MSA's text model, and, as a result, EvoMSA's first model is $g \circ m_b$. On the other hand, HA dataset is composed of texts and their associated polarity (negative, neutral, or positive), and, it is not related to TR. Consequently, it is feasible to create a text classifier that outputs the polarity of a given text. That is $m_2 = g_b \circ m_b$ where $m_b$ is B4MSA's text model (using the parameters shown in Table~\ref{tab:b4msa-parameters}) and $g_b$ is a linear SVM trained on HA, therefore EvoMSA's second model is $g \circ g_b \circ m_b$.

\subsection{Lexicon-based Model}

The text model, $m_3$, introduces external knowledge into our approach by the use of lexicons such as affective words. Thumbs Up-Down (TH) model, $m_3:$ text $\rightarrow \mathbb{R}^2$, counts the number of affective words keeping a separate record for the positive and negative words. We created a positive-negative lexicon based on several affective lexicons for English~\cite{Liu2017EnglishLexicon} and Spanish~\cite{Albornoz2012LanguageAnalysis,sidorov2012,Perez-Rosas2012LearningSpanish} and enriched with WordNet~\cite{Miller1995WordNet:English}. In the case of Arabic, we translate the English lexicon to Arabic language using Google translate, service employing Googletrans API~\cite{googletrans}.

\subsection{Emoji Space}
\label{sec:EmojiSpace}

Inspired by DeepMoji~\cite{Felbo2017UsingSarcasm}, we create the text model, $m_4: \text{text} \rightarrow \mathbb{R}^{64}$; the core idea is to predict what emoji would be the most probable one for a given text. For this purpose, we learn a B4MSA model per language using 3.2 million examples of the 64 most frequent emojis in each language. This dataset consists in $50,000$ examples per emoji extracted from our own collected tweets, that is, we filtered out these examples from (approximately) $2.0 \times 10^9$ Arabic tweets, $2.3 \times 10^9$ English tweets, and to $3.7 \times 10^9$ Spanish tweets. A few simple rules were followed to create the datasets: i) each example contains only one type of emoji to reduce the ambiguity among predictions, ii) all re-tweets were removed, iii) a uniform sample was chosen to avoid any seasonality effect. Finally, each selected tweet is transformed into a text and emoji pair, where the emoji is the one in the text. All emojis were removed from the text while training. Consequently, the dataset is a supervised learning dataset. 

B4MSA uses this dataset to create the Emoji Space. Each text is transformed to the vector space defined by B4MSA's text model using Table~\ref{tab:b4msa-parameters} parameters with a one-vs-rest strategy to train the SVM (i.e., $m_4 = g_b \circ m_b$ where $m_b$ is B4MSA's text model and $g_b$ is a linear SVM). Instead of being interested in the most probable emoji given text, we are interested in the decision functions of all the classifiers given a text such that each coordinate represents an emoji. Consequently, a 64-dimension real-valued vector represents a text.

\begin{figure}[!h]
    \subfloat[Spanish\label{fig:emo-es}]{\includegraphics[width=0.32\columnwidth]{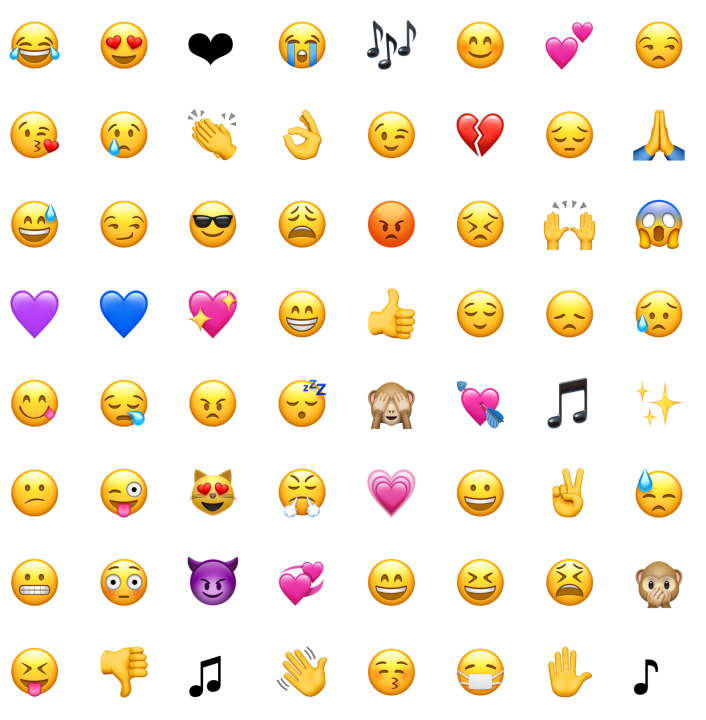}}\hfill
    \subfloat[English\label{fig:emo-en}]{\includegraphics[width=0.33\columnwidth]{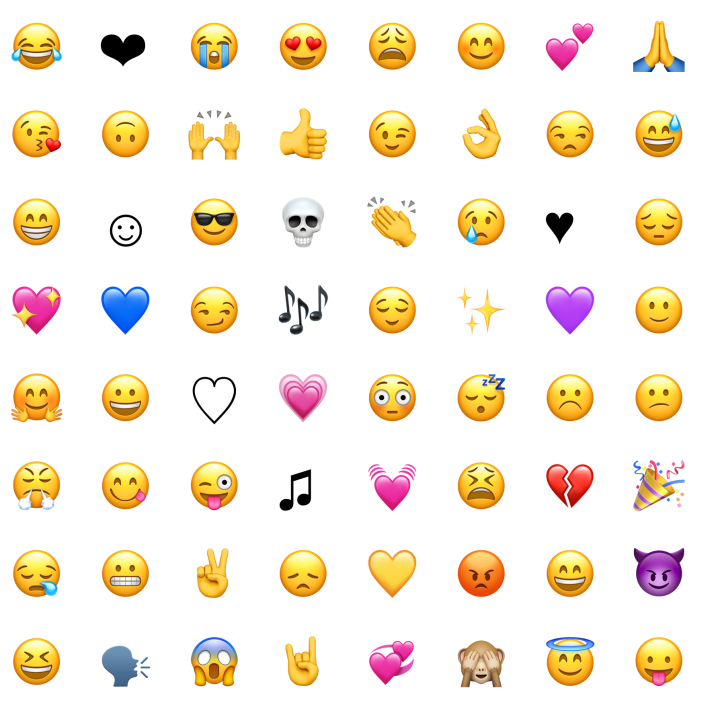}}\hfill
    \subfloat[Arabic\label{fig:emo-ar}]{\includegraphics[width=0.33\columnwidth]{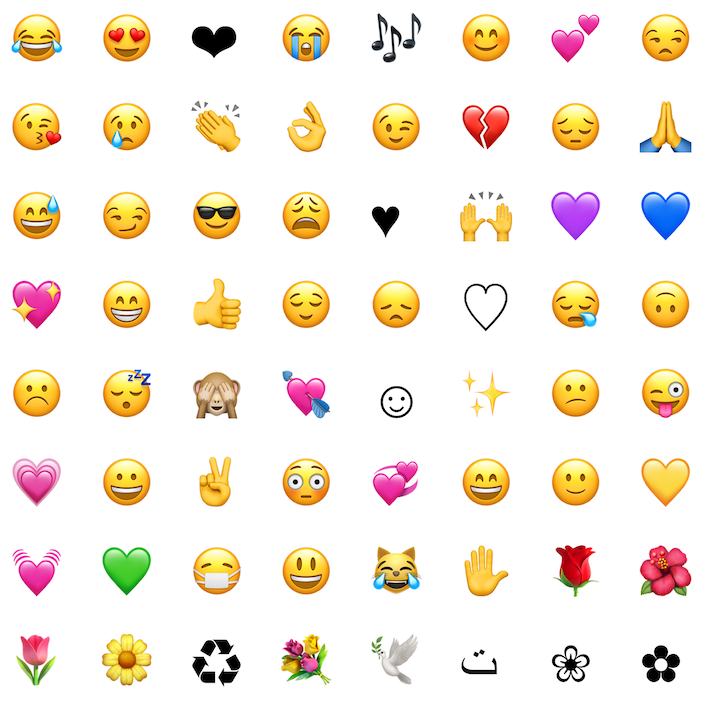}}

	\captionsetup{justification=centering}
    \caption{Actual emojis used by our Emoji-Space model.}
    \label{fig:emojis}
\end{figure}

Figure \ref{fig:emojis} lists the emojis used to create our Emoji-Space for Spanish, English, and Arabic languages; which also correspond to 64 most frequent emojis in these languages. The emojis are ordered row-wise being the most frequent the emoji in the left upper corner.
Notice the significant coincidence among the most frequent emojis in all languages.

\subsection{FastText}
\label{sec:FastText}

FastText~\cite{Bojanowski2016EnrichingInformation} is a tool to create text classifiers and learn a semantic vocabulary from a given collection of documents; this vocabulary is represented with a collection of high dimensional vectors, one per word. FastText is robust to lexical errors supporting out-vocabulary words, and it is used to represent a text into a vector space using the pre-computed models (see~\cite{Grave2018LearningLanguages}) for Arabic, English, and Spanish. In particular, each text is transformed into a vector using the vector sentences flag; these are vectors in 300 dimensions using the default parameters (i.e., $m_5: \text{text} \rightarrow \mathbb{R}^{300}$). 

\subsection{EvoDAG}

EvoDAG\footnote{https://github.com/mgraffg/EvoDAG}~\cite{Graff2016,Graff2017} is a steady-state GP system with tournament selection (tournament size 2) specifically tailored to tackle classification and regression problems. GP is an evolutionary algorithm with the distinctive characteristic of searching in a program search space, in particular, in this contribution, GP searches in a search space, $\Omega$, of functions. That is, $\Omega$ is the set of functions created by recursively composing elements from two sets: function set $\mathcal F$, and terminal set $\mathcal L$. The function set is composed by operations such as sum, product, sin, cos, max, and min, among others; and the inputs compose the terminal set, and normally, by an ephemeral random constant. Nonetheless, EvoDAG's terminal set only contains inputs, and each function, in the function set, is associated with a set of parameters that are identified using the training set. For example, let $f \in \mathcal F$ be a function of cardinality $1$ then $f(x \mid \theta) \in \Omega$ is an element of the search space, and, $\theta$ is identified with the training set using ordinary least squares, e.g., $f(x \mid \theta) = \theta \sin(x)$. 

In more detail, EvoDAG's search space is as follows: let $\mathcal F_c \subseteq \mathcal F$ be the functions with cardinality $c$ in the function set, and $\Omega^i$ be the elements created at iteration $i$, starting from $i=0$. Using this notation, the first elements, i.e., $i=0$, are $\Omega^0 = \{ \theta x_j \mid x_j \in \mathcal L \} \cup \bigcup_c \{ f(x_1, \cdots, x_c \mid \theta) \mid x_j \in \mathcal{L}, f \in \mathcal{F}_c \}$. The rest of the elements are composed recursively using $\Omega^i = \bigcup_c \{ f(x_1, \cdots, x_c \mid \theta) \mid x_j \in \Omega^{i-1}, f \in \mathcal{F}_c \}$; consequently, the search space is defined as $\Omega = \bigcup_i \Omega^i$. Using this notation, it is difficult to indicate that in the case of a commutative operator it is only included one of them, e.g., $x_i + x_j$ is included in the search space and $x_j + x_i$ it is not. The second restriction is that some functions require unique arguments. A function is decided to require unique arguments when $f(a, b, b, d) = \theta f(a, b, d)$ such as: $\min, \max$, and addition, among others.

EvoDAG searches $\Omega$ using a similar procedure than the one used to describe it. It is not possible to test all the elements at $\Omega^i$; instead, $\Omega^i$ is sampled, storing the elements in population $\mathcal P$. The initial population, $P^0$, contains $\{\theta x \mid x \in \mathcal L\}$, a set of functions such as Nearest Centroid Classifier, and other elements that are selected using the following procedure. A function, $f$, is randomly selected from $\mathcal F$, $f$'s arguments are randomly taken from $\mathcal L$ without replacement. This process continues until either all the elements of $\mathcal L$ have been selected, or the population size has been reached. In the former case, the process is to add one element at a time to $P^0$, choosing $f \in \mathcal F$ with the difference that $f$'s arguments are randomly taken from $P^0$; this mechanism continues until the population size is reached. For example, let $\mathcal F = \{+, \sin \}$ and $\mathcal L =\{x_1, x_2, x_3\}$ then $\mathcal P^0$ starts with $\{\theta_1 x_1, \theta_2 x_2\}$, this is followed by selecting a function, assume $+$ is selected, $P^0$ is $\{\theta_1 x_1, \theta_2 x_2, \theta_3 x_1 + \theta_4 x_2\}$, assume the next function selected is $\sin$, consequently, $P^0=\{\theta_1 x_1, \theta_2 x_2, \theta_3 x_1 + \theta_4 x_3, \theta_5 \sin{(x_2)}\}$. At this point all the inputs have been selected so the process continues by selecting the arguments from $\mathcal P^0$, suppose $\sin$ is selected and $\theta_3 x_1 + \theta_4 x_3$ is its argument, this makes $P^0=\{\theta_1 x_1, \theta_2 x_2, \theta_3 x_1 + \theta_4 x_3, \theta_5 \sin{(x_2)}, \theta_6 \sin{(\theta_3 x_1 + \theta_4 x_3)}\}$. This process is repeated until $\left| P^0 \right|$ reaches the population size.

Once the initial population is created, $P^0$, the evolution starts. EvoDAG uses a steady-state evolution, and, thus, it is not necessary to keep track of the population through the generations, therefore $\mathcal P=P^0$. The procedure used in the first generation is to create an element by selecting a function from $\mathcal F$, and its arguments are randomly selected from $\mathcal P$. The element created replaces an element of $\mathcal P$ which is selected using a negative tournament selection. From the second generation to the end of the run an element is created by first selecting $f$ and its argument are selected using tournament selection on $\mathcal P$, the element created replaces an element selected, from $\mathcal P$, with a negative tournament. 

Traditionally in GP, the evolution stops when the maximum number of generations is reached, or the fitness reaches a particular value; however, EvoDAG uses an early stopping approach. That is, the training set is split into a smaller training set, used to identify $\theta$ and the fitness of the individuals, and a validation set. Then, the best element is the one with the best performance on the validation set. The evolution stops when the best individual has not been updated in some evaluations, $4000$ is the default. 

EvoDAG function set is $\mathcal F = \{ \sum_{60},$ atan, $\text{NC}_{2}, \left| \cdot \right|, $hypot, $\max_5, \min_5, \prod_{20}, \text{NB}_5, \text{MN}_5, \sin, \sqrt{~}, \tan, \tanh \}$. Let us start by describing the addition which is defined as $f(x_1, \ldots, x_{60})=\sum_i \theta_i x_i$ where coefficients $\theta$ are identified with ordinary least squares (OLS) using the training set. Functions such as $\min$ and trigonometric functions are defined as $\theta f(x_i,\ldots)$ where $\theta$ is identified using OLS. For classification problems, the technique one-vs-rest is used, producing $k$ binary classification problems, and, for each different coefficients are optimized. $\text{NC}_5$ is the nearest centroid classifier whose output is the distance to each class centroid. $\text{NB}_5$ and $\text{MN}_5$ are Naive Bayes classifiers using Gaussian and Multinomial distributions, respectively. The outputs of these classifiers are the log-likelihood. NC, NB, and MN are always included in the initial population $P^0$ having as arguments all the inputs, in the rest of the run, these functions use their default number of arguments.

\begin{figure}[h]
\centering
    \includegraphics[width=0.60\textwidth]{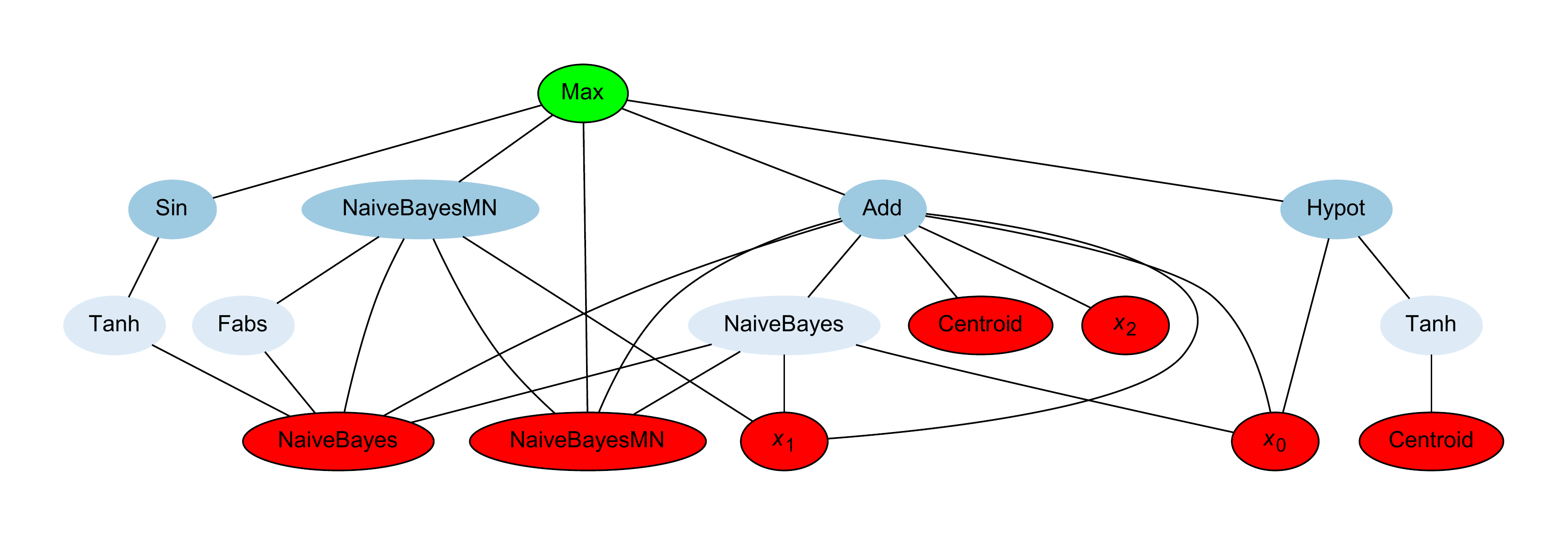}
\caption{An example of an evolved model by EvoDAG for the Arabic sentiment analysis task.}
\label{fig:evodag-model}
\end{figure}

Figure~\ref{fig:evodag-model} depicts a model evolved for the Arabic sentiment analysis task. As can be seen, the model is represented using a Direct Acyclic Graph (DAG) where the direction of the edges and dependency is bottom-up, e.g., $\tanh$ depends on Centroid, i.e., the hyperbolic tangent function is applied to Centroid's output. The input nodes are colored in red, the internal nodes are blue (the intensity is related to the distance to the height, the darker, the closer), and the green node is the output node. As mentioned previously, EvoDAG uses as inputs the decision functions of the models, the first three inputs (i.e., $X_0$, $X_1$, and $X_2$) correspond to the decision function values of the negative, neutral, and positive polarity of B4MSA model, the rest of the red nodes correspond to functions that are always in the initial population. It is important to mention that EvoDAG does not have information regarding whether input $X_i$ comes from a particular polarity decision function, consequently from EvoDAG point of view all inputs are equivalent.

%%%%%%%%%%%%%%%%%%%%%%%%%%%%%%%%%%%%%%%%%%%%%%%%%%%%%%%%%%%%%%%%%%%%%%%%%%%%%%%%%%%%%%%%%%%%%%%

\section{Competition, and Human Annotated Datasets}
\label{sec:datasets}

As mentioned before, we use the human annotated (HA) datasets from \cite{multilingual2016,CorpusArabic:2015} for Arabic, English, and Spanish languages. These are polarity datasets with three classes positive, neutral and negative, containing $223,306$ tweets for Spanish, $73,240$ for English, and $1,972$ for Arabic.

On the other hand, the datasets used to analyze EvoMSA's performance are described in Table~\ref{tab:number_tweets}.  These datasets include the competitions SemEval 2017 \cite{Rosenthal2017SemEval-2017Twitter} and 2018 \cite{Mohammad2018SemEval-2018Tweets}, TASS 2017 \cite{Martinez-Camara2017Overview2017} and 2018 \cite{Martinez-Camara2018OverviewEmotions}, HAHA 2018 \cite{Castro2018Overview2018} and MEX-A3T 2018 \cite{Alvarez-Carmona2018OverviewTweets}. It is worth to mention that for corpus InterTASS (TASS 2018) and MEX-A3T, we do not have the gold standard used in the competition; so, we performed cross-validation instead. Therefore, the performances reported are on that cross-validation dataset, and cannot be compared with the official performance presented by the competition.

These competitions present different tasks starting from the traditional sentiment analysis which corresponds to identify the polarity of a text; moving on to emotion ordinary classification where the emotions considered are anger, fear, joy, and sadness; safe-unsafe news classification; humor and aggressiveness detection. The majority of the problems are multi-class problems, and there are three binary classification problems which are safe news, humor, and aggressive detection. 

\begin{table}[!ht]
  \centering
  \caption{Number of tweets on the training and test sets in different tasks regarding sentiment analysis (SA), emotion-ordinal classification (EC), safe-unsafe classification of news (SUS), humor analysis (HA), and aggressive analysis (AA).}
  \begin{tabular}{cl rr}
\toprule
Language &  & Training & Test \\
\midrule
\multicolumn{4}{c}{SemEval 2017 (SA)} \\ \midrule
English & & 50,333 & 12,284 \\
Arabic  & & 3,355 & 6,100 \\
\midrule
\multicolumn{4}{c}{SemEval 2018 (EC)} \\ \midrule
		& anger	& 1,027	& 373 \\
		& fear	& 1,028	& 372 \\
Arabic	& joy	    & 952	& 448 \\
		& sadness	& 1,030	& 370 \\
		& valence	& 1,070	& 730 \\
\cmidrule{2-4}
		& anger		& 2,089	& 1,002 \\
        & fear		& 2,641	& 986 \\
English & joy		& 1,906	& 1,105 \\
        & sadness	& 1,930	& 975 \\
        & valence	& 1,630	& 937 \\
\cmidrule{2-4}
		& anger		& 1,359	& 627 \\
		& fear		& 1,368	& 618 \\
Spanish & joy		& 1,260	& 730 \\
		& sadness	& 1,350	& 641 \\
		& valence	& 1,795	& 648 \\
\midrule
% \multicolumn{4}{c}{TASS 2017 (SA)} \\ \midrule
\multicolumn{4}{c}{Spanish} \\ \midrule
\raisebox{0pt}[0pt][0pt]{
\raisebox{-1.5ex}{TASS 2017 (SA)}} & G. Corpus & 7,219 & 60,798 \\ %\cmidrule{2-4}
		& InterTASS &  1,514 & 1,899 \\
\cmidrule{2-4}
% \multicolumn{4}{c}{TASS 2018 (SUS)} \\ \midrule
		        & S1-L1 & 1,500 & 500 \\
TASS 2018 (SUS) & S1-L2 & 1,500 & 13,152 \\
		        & S2    & 274  & 407 \\ % \cmidrule{2-4}
% \midrule
% \multicolumn{4}{c}{HAHA 2018 (HA)} \\ \midrule
\cmidrule{2-4}
HAHA 2018 (HA) & & 16,000 & 4,000\\
\cmidrule{2-4}
% \multicolumn{4}{c}{MEX-A3T 2018 (AA)} \\ \midrule
MEX-A3T (AA)    &  & 5,389 & 2,311\\
\bottomrule
  \end{tabular}
  \label{tab:number_tweets}
\end{table}

%%%%%%%%%%%%%%%%%%%%%%%%%%%%%%%%%%%%%%%%%%%%%%%%%%%%%%%%%%%%%%%%%%%%%%%%%%%%%%%%%%%%%%%%%%%%%%%

\section{Analysis}
\label{sec:results}

This section presents EvoMSA's performance using the models described in Section~\ref{sec:description}, and on different competitions. The different EvoMSA instances are a combination of B4MSA trained with TR, B4MSA trained with HA, the Lexicon-based model (TH), Emoji Space (Emo), and FastText (FT). In total, there are 31 different combinations of these models; however, we decided to present only those combinations that had a significant impact on performance following a bottom-up approach. The starting point is EvoMSA using only TR, and then the remaining models are incorporated and tested one at a time. The model pair with the best performance is kept, and, the process continues testing the remaining text models until all of them are incorporated into EvoMSA. 

\begin{figure}[!htb]
\centering
\includegraphics[width=0.40\textwidth]{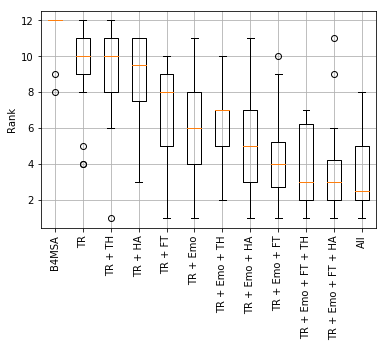}
\caption{Boxplot of the ranks of B4MSA and EvoMSA using different models, namely trained with the training set (TR), Lexicon-based model (TR + TH),  the human-annotated dataset (TR + HA), the FastText model (TR + FT), and the combination of these model until all the text model are combined with EvoMSA (All).}
\label{fig:models-rank}
\end{figure}

Figure~\ref{fig:models-rank} presents a boxplot of the ranks of EvoMSA instances as well as the rank of B4MSA. These ranks were calculated using all the datasets used, and, the performance measures used in each particular competition, namely macro-F1, macro-Recall, and Pearson correlation. From the figure, it can be observed that B4MSA has the highest rank, followed by EvoMSA using only TR. The difference in performance between these two systems is statistically significant with the confidence of 95\%, using Wilcoxon signed-rank test \cite{Wilcoxon1945IndividualMethods}. Comparing the performance of EvoMSA using only two models, it is observed that the Emoji Space (TR + Emo) is the one with the lowest rank, followed by FastText (TR + FT), the human-annotated dataset (TR + HA), and the Lexicon-based model. This latter model has an equivalent rank that EvoMSA using TR, albeit, it presented an outlier obtaining the best performance in one problem. The combination of TR, Emo, and FT has the lowest rank among the systems with three models, and EvoMSA with fourth (TR + Emo + FT + TH and TR + Emo + FT +HA) and five models (All) obtained similar ranks, having the lowest rank EvoMSA with all the models. In order to complement this boxplot, it was performed a comparison between EvoMSA with all the models (system with the lowest rank) and the rest of the systems. The statistical test used was a Wilcoxon signed-rank test \cite{Wilcoxon1945IndividualMethods}, and the $p$-values were adjusted with the Holm-Bonferroni method \cite{Holm1979AProcedure} to consider the multiple comparisons. The result is that the difference in performance between the best system and the next three best-performing systems is not significant with confidence of 95\%, whereas it is statistically different with the remaining of the systems. 

\begin{table}[!ht]
\centering
\caption{Performance comparison, in terms of macro-Recall (SemEval 2017) and macro-F1 (TASS 2017), of EvoMSA (using all models) with teams that participated in each competition; we include the performance of our baseline (B4MSA) and our participating system (INGEOTEC). The best performance for each task is indicated in boldface.}
\begin{tabular}{rcc cc}
\toprule
       & \multicolumn{2}{c}{SemEval 2017~\cite{Rosenthal2017SemEval-2017Twitter}} & \multicolumn{2}{c}{TASS 2017 ~\cite{Martinez-Camara2017Overview2017} } \\ \cmidrule(r){2-3}\cmidrule(l){4-5}
Systems/Teams & Arabic & English & G. Corpus & InterTASS  \\
\midrule
BB\_twtr & - & $\mathbf{0.681}$ & - & - \\
DataStories & - & $\mathbf{0.681}$ & - & - \\
LIA & - & $0.676$ & - & - \\
Senti17 & - & $0.674$ & - & - \\
NileTMRG & $0.583$ & - & - & - \\
NNEMBs & - & $0.669$ & - & - \\
EvoMSA & $\mathbf{0.592}$ & $0.662$ & $0.540$ & $0.474$\\
Tweester & - & $0.659$ & - & - \\ 
INGEOTEC & $0.477$ & $0.649$ & $\mathbf{0.577}$ & $\mathbf{0.507}$ \\
SiTAKA & $0.550$ &  $0.645$ & - & - \\
% HLP@UPENN(abeed)\cite{sarker-gonzalez:2017:SemEval} & $0.415$ & $0.637$ & - & - \\
ELiRF-UPV & $0.478$ & $0.619$ & $0.549$ & - \\
B4MSA & $0.510$ & $0.610$ & $0.490$ & $0.388$\\
% LSIS(amalhtait)\cite{htait-fournier-bellot:2017:SemEval} & $0.469$ & $0.561$ & - & - \\
% ONAM(ramybaly)\cite{baly-EtAl:2017:SemEval} & $0.422$ & $0.542$ & - & - \\
jacerong & - & - & $0.569$ & - \\
% ELiRF-UPV\cite{Hurtado2017ELiRF-UPVLearning} & - & - & $0.549$ & - \\
RETUYT & - & - & $0.546$ & - \\
tecnolengua & - & - & $0.528$ & - \\

\bottomrule
\end{tabular}
\label{tab:performance2017}
\end{table}

Table~\ref{tab:performance2017} presents the performance of EvoMSA using all the models, B4MSA, our participating system (i.e., INGEOTEC), and a selection of systems that participated in SemEval 2017 \cite{Rosenthal2017SemEval-2017Twitter} and TASS 2017 \cite{Martinez-Camara2017Overview2017}. Given that more than 30 teams participated in SemEval 2017, we decided to include only those systems that outperformed INGEOTEC in any of the languages. Regarding TASS 2017, the teams selected are the best submission of each team, and, those that obtained better performance than B4MSA which is our baseline. The performance in English sorts all systems. Comparing the performance of EvoMSA against the other competitors in SemEval 2017, it is observed that EvoMSA would have obtained the first place in Arabic and the sixth position in English. Regarding General Corpus (TASS 2017), our INGEOTEC team obtained the best performance, and EvoMSA would have been in the fifth place \cite{Martinez-Camara2017Overview2017}. 

Let us move our attention to those teams that participated in more than one language, the table presents only two out of three teams that participated in both languages, namely SiTAKA \cite{Jabreel2017SiTAKAFeatures} and ELiRF-UPV \cite{Gonzalez2017ELiRF-UPVLearning,Hurtado2017ELiRF-UPVLearning}; it can be observed that EvoMSA obtained the best performance among these teams, and, in addition only SiTAKA is better than B4MSA (our baseline) in both languages. On the other hand, ELiRF-UPV participated in both languages and competitions. This team had better performance than EvoMSA in TASS 2017 and worst in Arabic and English. 

\begin{table}[!htp]
\centering
\caption{Performance comparison in terms of Pearson correlation of EvoMSA (using all models) and teams that participated in SemEval 2018; we also included our baseline (B4MSA) and our participating system (INGEOTEC) in the listing. The best performance for each task is indicated in boldface.}
\begin{tabular}{rccccc}
\toprule
Systems/Teams & Anger & Fear & Joy & Sadness & Valence \\
\midrule
&\multicolumn{5}{c}{Arabic} \\ \cmidrule{2-6}
EiTAKA & $\mathbf{0.572}$ & $0.529$ & $\mathbf{0.634}$ & $0.563$ &  $\mathbf{0.809}$ \\
EvoMSA & $0.492$ & $0.495$ & $0.612$ & $\mathbf{0.622}$ & $0.761$ \\
% EvoMSA & $0.496$ & $0.527$ & $0.630$ & $0.610$ & $0.768$\\
AffectThor & $0.551$ & $\mathbf{0.551}$ & $0.631$ & $0.618$ & $0.752$ \\
INGEOTEC & $0.387$ & $0.440$ & $0.498$ & $0.425$ & $0.749$ \\
UNCC & $0.459$ & $0.483$ & $0.538$ & $0.587$ & $0.748$ \\
B4MSA & $0.425$ & $0.409$ & $0.401$ & $0.480$ & $0.680$\\
UWB~\cite{Priban2018UWBTweets} & $0.327$ & $0.345$ & $0.437$ & $0.467$ & - \\
EMA~\cite{badaro-EtAl:2018EMA}& $0.077$ &	$0.242$ & $0.215$ & $0.535$ & $0.643$ \\
% NileTMRG no tiene paper
NileTMRG~\cite{Mohammad2018SemEval-2018Tweets} & - & - & - & -  & $0.622$ \\
% LTL\_DUE / LTL Uni-Due & $0.301$ & $0.242$ & $0.360$ & $0.315$ & $0.552$ \\
\midrule
&\multicolumn{5}{c}{English} \\ \cmidrule{2-6}

SeerNet & $\mathbf{0.706}$ & $\mathbf{0.637}$ & $\mathbf{0.720}$ & $\mathbf{0.717}$ & $\mathbf{0.836}$ \\
PlusEmo2Vec & $0.704$ & $0.528$ & $\mathbf{0.720}$ & $0.683$ & $0.833$ \\
Amobee~\cite{rozental-fleischer:2018Amobee} & $0.667$ & $0.536$ & $0.705$ & $0.673$ & $0.813$ \\
psyML~\cite{gee-wang:2018psyML} & $0.670$ & $0.588$ & $0.686$ & $0.667$ & $0.802$ \\
EiTAKA~\cite{Jabreel2018EiTAKATweets} & $0.651$ & $0.595$ & $0.651$ & $0.636$ & $0.796$ \\
FOI DSS~\cite{karasalo-EtAl:2018FOIDSS} & $0.631$ & $0.521$ & $0.617$ & $0.591$ & $0.777$ \\
TCS Research~\cite{meisheri-dey:2018TCSResearch} & $0.641$ & $0.561$ & $0.655$ & $0.621$ & $0.777$ \\
NTUA-SLP~\cite{baziotis-EtAl:2018NTUA-SLP} & $0.644$ & $0.581$ & $0.678$ & $0.643$ & $0.777$ \\
AffecThor~\cite{Abdou2018AffecThorTweets} & $0.620$ & $0.538$ & $0.686$ & $0.622$ & $0.776$ \\
Epita~\cite{davalfrerot-bouchekif-moreau:2018Epita} & - & - & - & - & $0.776$ \\
INGEOTEC & $0.560$ & $0.489$ & $0.643$ & $0.584$ & $0.760$ \\
ELiRF-UPV~\cite{Gonzalez2018ELiRF-UPVTweets} & $0.601$ & $0.525$ & $0.630$ & $0.605$ & $0.759$\\
EvoMSA & $0.575$ & $0.520$ & $0.656$ & $0.605$ & $0.753$\\
% EvoMSA & $0.578$ & $0.524$ & $0.662$ & $0.602$ & $0.755$\\
UNCC~\cite{Abdullah2018TeamUNCCLearning} & $0.604$ & $0.544$ & $0.638$ & $0.610$ & $0.736$ \\
YNU-HPCC~\cite{zhang-wang-zhang:2018YNU-HPCC} & $0.554$ & $0.523$ & $0.624$ & $0.610$ & $0.733$ \\
B4MSA & $0.420$ & $0.400$ & $0.521$ & $0.487$ & $0.530$\\
% LTL\_DUE / LTL Uni-Due & $0.401$ & $0.078$ & $0.468$ & $0.384$ & $0.491$ \\
% ZMU no tiene paper
ZMU~\cite{Mohammad2018SemEval-2018Tweets} & $0.556$ & $0.565$ & $0.586$ & $0.579$ & $0.122$ \\
\midrule
&\multicolumn{5}{c}{Spanish} \\
\cmidrule{2-6}
Amobee & - & - & - & - & $\mathbf{0.765}$ \\
EvoMSA & $0.560$ & $0.659$ & $\mathbf{0.693}$ & $0.672$ & $0.757$\\
AffectThor & $\mathbf{0.606}$ & $\mathbf{0.706}$ & $0.667$ & $\mathbf{0.677}$ & $0.756$ \\
% EvoMSA & $0.546$ & $0.646$ & $0.677$ & $0.690$ & $0.744$\\
ELiRF-UPV~\cite{Gonzalez2018ELiRF-UPVTweets} & $0.520$ & $0.567$ & $0.592$ & $0.620$ & $0.729$ \\
INGEOTEC & $0.468$ & $0.634$ & $0.655$ & $0.628$ & $0.698$ \\
UG18~\cite{kuijper-vanlenthe-vannoord:2018UG18} & $0.499$ & $0.606$ & $0.665$ & $0.625$ & $0.682$ \\
YNU-HPCC~\cite{zhang-wang-zhang:2018YNU-HPCC} & $0.263$ & $0.283$ & $0.513$ & $0.380$ & $0.556$ \\
B4MSA & $0.454$ & $0.568$ & $0.570$ & $0.546$ & $0.538$\\
% LTL\_DUE / LTL Uni-Due & $0.203$ & $0.352$ & $0.412$ & $0.400$ & $0.477$ \\
UWB~\cite{Priban2018UWBTweets} & $0.361$ & $0.606$ & $0.544$ & $0.506$ & - \\
\bottomrule
\end{tabular}
\label{tab:semeval2018}
\end{table}

Table \ref{tab:semeval2018} shows the results achieved on SemEval 2018 \cite{Mohammad2018SemEval-2018Tweets} datasets. The table includes the performance of EvoMSA, B4MSA, INGEOTEC, and a selection of competitors. The teams included were those that obtained a better position than INGEOTEC in English and those that outperformed the competition baseline on Arabic and Spanish. The table is organized according to the competition language, namely Arabic, English, and Spanish. Furthermore, the systems are sorted by valence in all the languages. From the table, it can be observed that EvoMSA in Arabic would have obtained the second place in valence, first in sadness, and third in the rest of the tasks. On the other hand, in English, EvoMSA did not outperform INGEOTEC in valence; nonetheless, it did improve INGEOTEC in the rest of the problems. In Spanish, EvoMSA would have been in first place in joy and second place in the rest of the tasks. 

Seven teams participated in two or more languages; only AffectThor \cite{Abdou2018AffecThorTweets} submitted results for all languages. Their approach in English outperformed EvoMSA in all tasks; in Arabic and Spanish languages AffectThor obtained better performance in anger, fear, and joy. Three teams submitted results for Arabic and English: EiTAKA \cite{Jabreel2018EiTAKATweets} obtained better performance than EvoMSA in all tasks; UNCC \cite{Abdullah2018TeamUNCCLearning} in English had a better position than EvoMSA in anger, fear, and sadness.
EvoMSA outperforms UWB \cite{Priban2018UWBTweets} in all tasks. Finally, three teams participated in English and Spanish, Amobee \cite{rozental-fleischer:2018Amobee} obtained a better score than EvoMSA in all the tasks; ELiRF-UPV \cite{Gonzalez2018ELiRF-UPVTweets} obtained better results in English and worst in Spanish; and YNU-HPCC \cite{zhang-wang-zhang:2018YNU-HPCC} only outperformed EvoMSA in fear and sadness in English.

\begin{table}[!ht]
\centering
\caption{Performance in terms of macro-F1 (except HAHA where F1 is used instead) of EvoMSA (with all the models), our baseline (B4MSA), our participating system (INGEOTEC), and other competitors. The performance in S1-L1 sorts the systems and the best performance is indicated in boldface for each task.}
%\resizebox{\columnwidth}{!}
{
\begin{tabular}{l ccc cc}
\toprule
   & \multicolumn{3}{c}{TASS 2018~\cite{Martinez-Camara2018OverviewEmotions}} & \multicolumn{2}{c}{IberEval 2018 ~\cite{Castro2018Overview2018,Alvarez-Carmona2018OverviewTweets}}\\ \cmidrule(r){2-4} \cmidrule(l){5-6}
System & S1-L1 & S1-L2 & S2 & HAHA & MEX-A3T  \\ \midrule
INGEOTEC & $\mathbf{0.795}$ & $0.866$ & $0.719$ & $0.797$ & $\mathbf{0.794}$ \\
ELiRF & $0.790$ & $\mathbf{0.883}$ & $0.699$ & $0.772$ & - \\ 
EvoMSA & $0.783$ & $0.854$ & $\mathbf{0.725}$ & $\mathbf{0.799}$ & $0.789$\\
rbnUGR & $0.774$ & $0.873$ & $0.683$ & - & - \\
% EvoMSA & $0.774$ & $0.854$ & $\mathbf{0.723}$ & $0.797$ & $0.786$\\
M.CLOUD& $0.767$ & $0.793$ & $0.651$ & - & - \\
SINAI & $0.728$ & $0.773$ & - & - & - \\
B4MSA & $0.722$ & $0.768$ & $0.519$ & $0.793$ & $0.786$\\
% UO\_UPV & - & - & - & $0.7851$ & - \\
\bottomrule
\end{tabular}
}
\label{tab:SEPLN}
\end{table}

Table~\ref{tab:SEPLN} shows the performance of EvoMSA (using all models), B4MSA, and the participants of TASS 2018 \cite{Martinez-Camara2018OverviewEmotions} and IberEval 2018 \cite{Castro2018Overview2018}. The table shows that EvoMSA would have obtained two first places and it did not outperform our participating system, INGEOTEC on the rest of the tasks. 

\begin{figure}[!htb]
\centering
\includegraphics[width=0.40\textwidth]{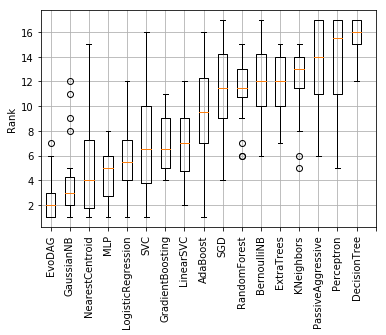}
\caption{Ranking comparison of different classifiers on all tested datasets.}
\label{fig:classifiers-rank}
\end{figure}

After analyzing the behavior of EvoMSA, it is time to measure the effects that EvoDAG, has in the overall performance. The procedure used is to replace EvoDAG in EvoMSA (using all models) by, almost, all the classifiers implemented in \cite{Pedregosa2011Scikit-learn:Python} with their default parameters. In total, sixteen different classifiers are used to perform this comparison. Figure~\ref{fig:classifiers-rank} presents the boxplot of the ranks of these classifiers as well as EvoDAG. The figure is ordered so that the classifier with the lowest rank is on the left and the classifier with the highest rank is on the right; this is only to facilitate the reading.  

It is observed from the figure that EvoDAG obtained the lowest rank (best performance), closely followed by a Gaussian Naive Bayes and the Nearest Centroid classifier, the Decision Trees presented the highest rank. EvoDAG, Gaussian NB, and Nearest Centroid obtained the first position in a number of problems; however, these are not the only ones obtaining the first position, these other classifiers are Logistic regression, SVM, and Ada Boost. The comparison between EvoDAG's performance against the other systems --using the Wilcoxon signed-rank test \cite{Wilcoxon1945IndividualMethods} and adjusting the $p$-values with Holm-Bonferroni method \cite{Holm1979AProcedure} to consider the multiple comparisons-- shows that there is a difference in performance with confidence of 95\%. Nonetheless, as mentioned, there are classifiers that in some problems obtained the best place. 

%%%%%%%%%%%%%%%%%%%%%%%%%%%%%%%%%%%%%%%%%%%%%%%%%%%%%%%%%%%%%%%%%%%%%%%%%%%%%%%%%%%%%%%%%%%%%%%

\section{Conclusions}
\label{sec:conclusions}

We presented EvoMSA, a multilingual and domain-independent sentiment analysis system. EvoMSA is designed to combine different resources into one objective; among the possible resources, we use domain-specific training set and other related human-annotated datasets, lexicon-based models, semi-supervised models like our Emoji Space and FastText. These models are combined into a classifier based on GP to produce the final prediction. EvoMSA's components are analyzed based on performance, including our classifier EvoDAG. The study shows that the resource contributing most to the performance is Emoji Space; on the other hand, the system with the lowest average rank (the lower, the better), was produced by using all the resources. Furthermore, it is worth to mention that replacing EvoDAG with a simpler classifier such as Gaussian Naive Bayes one can reduce computing time, nonetheless, with a performance impact.

EvoMSA performance is analyzed using datasets from different competitions, namely SemEval 2017 and 2018, TASS 2017 and 2018, HAHA 2018, and MEX-A3T 2018. It is important to note that almost all the parameters of EvoMSA and its components are kept constant in all the datasets. Consequently, EvoMSA can be considered an almost free parameter algorithm in a multilingual domain. Furthermore, the result shows that EvoMSA is competitive against the systems participating in those competitions. Based on our experimental results, EvoMSA would have obtained fifth first places (SemEval 2017 in Arabic; SemEval 2018 sadness in Arabic and joy in Spanish; TASS 2018 on S2 dataset; and HAHA 2018), on SemEval 2018 would be on average on the second place in Spanish and third in Arabic. These results are evidence that EvoMSA has a significant generalization potential over several languages.
    
Finally, we would like to discuss some research avenues briefly. We have tested EvoMSA on different sentiment analysis competitions; however, the scheme is general enough to tackle multi-modal problems such as combining images and texts. We also have tested two semantic resources, i.e., Emoji Space and FastText, in the future, it would be essential to develop and test other semantic representations. An important characteristic that has not been addressed in EvoMSA is that, currently, the models evolved are not intended to be understood. Given that EvoDAG is a GP system, it would be desired that the model evolved would be a white box. 

%\section*{Acknowledgment}



\begin{thebibliography}{10}

\bibitem{Liu2012AAnalysis}
Bing Liu and Lei Zhang.
\newblock {A survey of opinion mining and sentiment analysis}.
\newblock In {\em Mining Text Data}, pages 415--463. Springer US, Boston, MA,
  Jan. 2012.

\bibitem{Poria2016FusingContent}
Soujanya Poria, Erik Cambria, Newton Howard, Guang-Bin Huang, and Amir Hussain.
\newblock {Fusing audio, visual and textual clues for sentiment analysis from
  multimodal content}.
\newblock {\em Neurocomputing}, 174:50--59, Jan. 2016.

\bibitem{Ortis2018VisualImages}
Alessandro Ortis, Giovanni~M. Farinella, Giovanni Torrisi, and Sebastiano
  Battiato.
\newblock {Visual sentiment analysis based on objective text description of
  images}.
\newblock In {\em 2018 International Conference on Content-Based Multimedia
  Indexing (CBMI)}, pages 1--6. IEEE, Sept. 2018.

\bibitem{Cambria2016}
Erik Cambria.
\newblock {Affective computing and sentiment analysis}.
\newblock {\em IEEE Intelligent Systems}, 31(2):102--107, Mar. 2016.

\bibitem{Martinez-Camara2018OverviewEmotions}
Eugenio Mart\'{i}nez-C\'{a}mara, Yudivián Almeida-Cruz, Manuel Carlos
  D\'{i}az-Galiano, Suilan Est\'{e}vez-Velarde, Migueí~A Garc\'{i}a-Cumbreras,
  Manuel Garc\'{i}a-Vega, Yoan Guti\'{e}rrez, Arturo Montejo-R\'{a}ez, Andrés
  Montoyo, Rafael Mu\~{n}oz, Alejandro Piad-Morffis, and Julio
  Villena-Rom\'{a}n.
\newblock {Overview of TASS 2018: Opinions, health and emotions}.
\newblock {\em CEUR Workshop Proceedings}, 2172:13--27, Sept. 2018.

\bibitem{Martinez-Camara2017Overview2017}
Eugenio Mart\'{i}nez-C\'{a}mara, Manuel~C D\'{i}az-Galiano, Angel
  Garc\'{i}a-Cumbreras, Manuel Garc\'{i}a-Vega, and Julio Villena-Rom\'{a}n.
\newblock {Overview of TASS 2017}.
\newblock {\em CEUR Workshop Proceedings}, 1896:13--21, Sept. 2017.

\bibitem{Mohammad2018SemEval-2018Tweets}
Saif~M Mohammad, Felipe Bravo-Marquez, Mohammad Salameh, and Svetlana
  Kiritchenko.
\newblock {SemEval-2018 Task 1: Affect in tweets}.
\newblock In {\em Proc. of the 12th International Workshop on Semantic
  Evaluation}, pages 1--17. ACL, June 2018.

\bibitem{Rosenthal2017SemEval-2017Twitter}
Sara Rosenthal, Noura Farra, and Preslav Nakov.
\newblock {SemEval-2017 Task 4: Sentiment analysis in Twitter}.
\newblock In {\em Proc. of the 11th International Workshop on Semantic
  Evaluation}, pages 502--518. ACL, Aug. 2017.

\bibitem{Moctezuma2017ATASS17}
Daniela Moctezuma, Mario Graff, Sabino Miranda-Jim\'enez, Eric~S Tellez, Abel
  Coronado, Claudia~N S\'anchez, and José Ortiz-Bejar.
\newblock {A genetic programming approach to sentiment analysis for Twitter:
  TASS'17}.
\newblock {\em CEUR Workshop Proceedings}, 1896:23--28, Sept. 2017.

\bibitem{Moctezuma2018INGEOTECCompetition}
Daniela Moctezuma, José Ortiz-Bejar, Eric~S Tellez, Sabino
  Miranda-Jim\'{e}nez, and Mario Graff.
\newblock {INGEOTEC solution for Task 4 in TASS'18 competition}.
\newblock {\em CEUR Workshop Proceedings}, 2172:111--115, Sept. 2018.

\bibitem{Miranda-Jimenez2017INGEOTECAnalysis}
Sabino Miranda-Jim\'{e}nez, Mario Graff, Eric~S Tellez, and Daniela Moctezuma.
\newblock {INGEOTEC at SemEval 2017 Task 4: A B4MSA ensemble based on genetic
  programming for Twitter sentiment analysis}.
\newblock In {\em Proc. of the 11th International Workshop on Semantic
  Evaluation}, pages 771--776. ACL, Aug. 2017.

\bibitem{Graff2018}
Mario Graff, Sabino Miranda-Jim\'{e}nez, Eric~S Tellez, and Daniela Moctezuma.
\newblock {INGEOTEC at SemEval-2018 Task 1: EvoMSA and microTC for sentiment
  analysis}.
\newblock In {\em Proc. of the 12th International Workshop on Semantic
  Evaluation}, pages 146--150, June 2018.

\bibitem{Graff2018a}
Mario Graff, Sabino Miranda-Jim\'{e}nez, Eric~S Tellez, Daniela Moctezuma,
  Vladimir Salgado, Jos\'e Ortiz-Bejar, and Claudia~N S\'{a}nchez.
\newblock {INGEOTEC at MEX-A3T: Author profiling and aggressiveness analysis in
  Twitter using microTC and EvoMSA}.
\newblock {\em CEUR Workshop Proceedings}, 2150:128--133, Sept. 2018.

\bibitem{Ortiz-Bejar2018}
Jose Ortiz-Bejar, Vladimir Salgado, Mario Graff, Daniela Moctezuma, Sabino
  Miranda-Jimenez, and Eric~Sadit Tellez.
\newblock {INGEOTEC at IberEval 2018 Task HaHa: MicroTC and EvoMSA to detect
  and score humor in texts}.
\newblock {\em CEUR Workshop Proceedings}, 2150:195--202, Sept. 2018.

\bibitem{Felbo2017UsingSarcasm}
Bjarke Felbo, Alan Mislove, Anders S{\o}gaard, Iyad Rahwan, and Sune Lehmann.
\newblock {Using millions of emoji occurrences to learn any-domain
  representations for detecting sentiment, emotion and sarcasm}.
\newblock In {\em Proc. of the 2017 Conference on Empirical Methods in Natural
  Language Processing}, page 1615–1625, Copenhagen, Denmark, Sept. 2017. ACL.

\bibitem{Lo2017499}
S.~L. Lo, E.~Cambria, R.~Chiong, and D.~Cornforth.
\newblock {Multilingual sentiment analysis: from formal to informal and scarce
  resource languages}.
\newblock {\em Artificial Intelligence Review}, 48(4):499--527, Aug. 2017.

\bibitem{vilares2017}
David Vilares, Miguel~A Alonso, and Carlos G\'{o}mez-Rodr\'{i}guez.
\newblock {Supervised sentiment analysis in multilingual environments}.
\newblock {\em Information Processing {\&} Management}, 53(3):595--607, May
  2017.

\bibitem{Balahur2014}
Alexandra Balahur, Marco Turchi, Ralf Steinberger, Jose~Manuel Perea-Ortega,
  Guillaume Jacquet, Dilek Kucuk, Vanni Zavarella, and Adil~El Ghali.
\newblock {Resource creation and evaluation for multilingual sentiment analysis
  in social media texts}.
\newblock In {\em Proc. of the Ninth International Conference on Language
  Resources and Evaluation (LREC'14)}, pages 4265--4269, Reykjavik, Iceland,
  May 2014. European Language Resources Association (ELRA).

\bibitem{Meng:2012}
{Meng, Xinfan and Wei, Furu and Liu, Xiaohua and Zhou, Ming and Xu, Ge and
  Wang, Houfeng}.
\newblock {Cross-lingual mixture model for sentiment classification}.
\newblock In {\em Proc. of the 50th Annual Meeting of the Association for
  Computational Linguistics: Long Papers - Volume 1}, ACL '12, pages 572--581,
  Stroudsburg, PA, USA, July 2012. ACL.

\bibitem{becker2017}
Karin Becker, Viviane~P Moreira, and Aline G~L dos Santos.
\newblock {Multilingual emotion classification using supervised learning:
  Comparative experiments}.
\newblock {\em Information Processing {\&} Management}, 53(3):684--704, May
  2017.

\bibitem{ClicheBloomberg2017BBLSTMs}
Mathieu Cliche~Bloomberg.
\newblock {BB\_twtr at SemEval-2017 Task 4: Twitter sentiment analysis with
  CNNs and LSTMs}.
\newblock In {\em Proc. of the 11th International Workshop on Semantic
  Evaluations}, pages 573--580. ACL, Aug. 2017.

\bibitem{Baziotis2017DataStoriesAnalysis}
Christos Baziotis, Nikos Pelekis, and Christos Doulkeridis.
\newblock {DataStories at SemEval-2017 Task 4: Deep LSTM with attention for
  message-level and topic-based sentiment analysis}.
\newblock In {\em Proc. of the 11th International Workshop on Semantic
  Evaluations}, pages 747--754. ACL, Aug. 2017.

\bibitem{El-Beltagy2017NileTMRGAnalysis}
Samhaa~R El-Beltagy, Mona~El Kalamawy, and Abu~Bakr Soliman.
\newblock {NileTMRG at SemEval-2017 Task 4: Arabic sentiment analysis}.
\newblock In {\em Proc. of the 11th International Workshop on Semantic
  Evaluation}, pages 790--795. ACL, Aug. 2017.

\bibitem{Jabreel2017SiTAKAFeatures}
Mohammed Jabreel and Antonio Moreno.
\newblock {SiTAKA at SemEval-2017 Task 4: Sentiment analysis in Twitter based
  on a rich set of features}.
\newblock In {\em Proc. of the 11th International Workshop on Semantic
  Evaluation}, pages 694--699. ACL, Aug. 2017.

\bibitem{Gonzalez2017ELiRF-UPVLearning}
Jos\'e-A Gonz\'{a}lez, Ferran Pla, and Llu\'is-F Hurtado.
\newblock {ELiRF-UPV at SemEval-2017 Task 4: Sentiment analysis using deep
  learning}.
\newblock In {\em Proc. of the 11th International Workshop on Semantic
  Evaluation}, pages 723--727. ACL, Aug. 2017.

\bibitem{Zahran2015WordArabic}
Mohamed~A. Zahran, Ahmed Magooda, Ashraf~Y. Mahgoub, Hazem Raafat, Mohsen
  Rashwan, and Amir Atyia.
\newblock {Word representations in vector space and their applications for
  Arabic}.
\newblock In {\em Computational Linguistics and Intelligent Text Processing},
  pages 430--443. Springer, Cham, Apr. 2015.

\bibitem{Duppada2018SeerNetTweets}
Venkatesh Duppada, Royal Jain, and Sushant Hiray.
\newblock {SeerNet at SemEval-2018 Task 1: Domain adaptation for affect in
  tweets}.
\newblock {\em Proc. of the 12th International Workshop on Semantic
  Evaluation}, pages 18--23, June 2018.

\bibitem{Jabreel2018EiTAKATweets}
Mohammed Jabreel and Antonio Moreno.
\newblock {EiTAKA at SemEval-2018 Task 1: An ensemble of n-channels ConvNet and
  XGboost regressors for emotion analysis of tweets}.
\newblock In {\em Proc. of the 12th International Workshop on Semantic
  Evaluation}, pages 193--199. ACL, June 2018.

\bibitem{Abdou2018AffecThorTweets}
Mostafa Abdou, Artur Kulmizev, and Joan Gin\'{e}s~Ametll\'{e}.
\newblock {AffecThor at SemEval-2018 Task 1: A cross-linguistic approach to
  sentiment intensity quantification in tweets}.
\newblock In {\em Proc. of the 12th International Workshop on Semantic
  Evaluation}, pages 210--217. ACL, June 2018.

\bibitem{Hurtado2017ELiRF-UPVLearning}
Llu\'is-F Hurtado, Ferran Pla, and José-A Gonz\'{a}lez.
\newblock {ELiRF-UPV at TASS 2017: Sentiment analysis in Twitter based on deep
  learning}.
\newblock {\em CEUR Workshop Proceedings}, 1896:29--34, Sept. 2017.

\bibitem{Gonzalez2018:ELiRF-UPV}
Jos\'e-A Gonz\'{a}lez, Llu\'is-F Hurtado, and Ferran Pla.
\newblock {ELiRF-UPV at TASS 2018: Emotional categorization of news articles}.
\newblock {\em CEUR Workshop Proceedings}, 2172:103--109, Sept. 2018.

\bibitem{Alvarez-Carmona2018OverviewTweets}
Miguel~A Alvarez-Carmona, Estefanía Guzm\'{a}n-Falc\'{o}n, Manuel
  Montes-Y-G\'{o}mez, Hugo~Jair Escalante, Luis Villase\~{n}or Pineda,
  Verónica Reyes-Meza, and Antonio Rico-Sulayes.
\newblock {Overview of MEX-A3T at IberEval 2018: Authorship and aggressiveness
  analysis in Mexican Spanish tweets}.
\newblock {\em CEUR Workshop Proceedings}, 2150:74--96, Sept. 2018.

\bibitem{Castro2018Overview2018}
Santiago Castro, Luis Chiruzzo, and Aiala Ros\'{a}.
\newblock {Overview of the HAHA Task: Humor analysis based on human annotation
  at IberEval 2018}.
\newblock {\em CEUR Workshop Proceedings}, 2150:187--194, Sept. 2018.

\bibitem{EnriqueMunizCuza2018AttentionDetection}
Carlos Enrique Mu\~{n}iz Cuza, Gretel Liz De la Pe\~{n}a Sarrac\'{e}n, and
  Paolo Rosso.
\newblock {Attention mechanism for aggressive detection}.
\newblock In {\em CEUR Workshop Proceedings}, volume 2150, pages 114--118,
  Sept. 2018.

\bibitem{Aragon2018Author2018}
Mario~Ezra Arag\'{o}n and A~Pastor L\'{o}pez-Monroy.
\newblock {Author profiling and aggressiveness detection in Spanish tweets:
  MEX-A3T 2018}.
\newblock {\em CEUR Workshop Proceedings}, 2150:134--139, Sept. 2018.

\bibitem{Ortega-Bueno2018UOMedia}
Reynier Ortega-Bueno, Carlos~E Mu\~{n}iz Cuza, José~E Medina~Pagola, and Paolo
  Rosso.
\newblock {UO UPV: Deep linguistic humor detection in Spanish social media}.
\newblock {\em CEUR Workshop Proceedings}, 2150:203--213, Sept. 2018.

\bibitem{Joshi2017AutomaticDetection}
Aditya Joshi, Pushpak Bhattacharyya, and Mark~J. Carman.
\newblock {Automatic sarcasm detection}.
\newblock {\em ACM Computing Surveys}, 50(5):1--22, Sept. 2017.

\bibitem{Joshi2016AreDetection}
Aditya Joshi, Vaibhav Tripathi, Kevin Patel, Pushpak Bhattacharyya, and Mark
  Carman.
\newblock {Are word embedding-based features useful for sarcasm detection?}
\newblock In {\em Proc. of the 2016 Conference on Empirical Methods in Natural
  Language Processing}, pages 1006--1011, Stroudsburg, PA, USA, Nov. 2016. ACL.

\bibitem{Amir2016ModellingMedia}
Silvio Amir, Byron~C. Wallace, Hao Lyu, Paula Carvalho, and Mario~J. Silva.
\newblock {Modelling context with user embeddings for sarcasm detection in
  social media}.
\newblock In {\em Proc. of The 20th SIGNLL Conference on Computational Natural
  Language Learning}, pages 167--177, Stroudsburg, PA, USA, Aug. 2016. ACL.

\bibitem{Ghosh2016FrackingNetwork}
Aniruddha Ghosh and Dr.~Tony Veale.
\newblock {Fracking sarcasm using neural network}.
\newblock In {\em Proc. of the 7th Workshop on Computational Approaches to
  Subjectivity, Sentiment and Social Media Analysis}, pages 161--169,
  Stroudsburg, PA, USA, June 2016. ACL.

\bibitem{Wolpert1992StackedGeneralization}
David~H. Wolpert.
\newblock {Stacked generalization}.
\newblock {\em Neural Networks}, 5(2):241--259, Jan. 1992.

\bibitem{Graff2016}
Mario Graff, Eric~S. Tellez, Sabino Miranda-Jimenez, and Hugo~Jair Escalante.
\newblock {EvoDAG: A semantic genetic programming Python library}.
\newblock In {\em 2016 IEEE International Autumn Meeting on Power, Electronics
  and Computing (ROPEC)}, pages 1--6. IEEE, Nov. 2016.

\bibitem{Graff2017}
Mario Graff, Eric~S. Tellez, Hugo Jair~Escalante, and Sabino
  Miranda-Jim\'{e}nez.
\newblock {Semantic genetic programming for sentiment analysis}.
\newblock In Oliver Sch\"{u}tze, Leonardo Trujillo, Pierrick Legrand, and
  Yazmin Maldonado, editors, {\em NEO 2015}, pages 43--65. Springer, Cham, Aug.
  2017.

\bibitem{Tellez2017ATwitter}
Eric~S. Tellez, Sabino Miranda-Jim\'{e}nez, Mario Graff, Daniela Moctezuma,
  Ranyart~R. Su\'{a}rez, and Oscar~S. Siordia.
\newblock {A simple approach to multilingual polarity classification in
  Twitter}.
\newblock {\em Pattern Recognition Letters}, 94:68--74, July 2017.

\bibitem{Tellez2017AAnalysis}
Eric~S. Tellez, Sabino Miranda-Jim\'{e}nez, Mario Graff, Daniela Moctezuma,
  Oscar~S. Siordia, and Elio~A. Villase\~{n}or.
\newblock {A case study of Spanish text transformations for Twitter sentiment
  analysis}.
\newblock {\em Expert Systems with Applications}, 81:457--471, Sept. 2017.

\bibitem{Liu2017EnglishLexicon}
Bing Liu.
\newblock {English opinion lexicon}.
\newblock
  \url{https://www.cs.uic.edu/~liub/FBS/sentiment-analysis.html#lexicon}, 2017.
\newblock Accessed 26-Jul-2018.

\bibitem{Albornoz2012LanguageAnalysis}
Jorge Carrillo~De Albornoz, Laura Plaza, and Pablo Gerv.
\newblock {SentiSense: An easily scalable concept-based affective lexicon for
  sentiment analysis}.
\newblock In {\em International Conference on Language Resources and
  Evaluation}, pages 3562--3567, Istanbul, Turkey, 2012. European Language
  Resources Association (ELRA).

\bibitem{sidorov2012}
Grigori Sidorov, Sabino Miranda-Jim\'{e}nez, Francisco Viveros-Jim\'{e}nez,
  Alexander Gelbukh, Noé Castro-S\'{a}nchez, Francisco Vel\'{a}squez, Ismael
  D\'{i}az-Rangel, Sergio Su\'{a}rez-Guerra, Alejandro Trevi\~{n}o, and Juan
  Gordon.
\newblock {Empirical study of machine learning based approach for opinion
  mining in tweets}.
\newblock In {\em Proc. of the 11th Mexican International Conference on
  Advances in Artificial Intelligence - Volume Part I}, MICAI'12, pages 1--14,
  Berlin, Heidelberg, Oct. 2013. Springer-Verlag.

\bibitem{Perez-Rosas2012LearningSpanish}
Veronica Perez-Rosas, Carmen Banea, and Rada Mihalcea.
\newblock {Learning sentiment lexicons in Spanish}.
\newblock In {\em Proc. of the Eighth International Conference on Language
  Resources and Evaluation (LREC-2012)}, pages 3077--3081, May 2012.

\bibitem{Miller1995WordNet:English}
George~A. Miller and George A.
\newblock {WordNet: a lexical database for English}.
\newblock {\em Communications of the ACM}, 38(11):39--41, Nov. 1995.

\bibitem{googletrans}
googletrans.
\newblock {Googletrans: Free and unlimited Google translate API for Python}.
\newblock \url{http://py-googletrans.readthedocs.io/en/latest}, 2018.
\newblock Accessed 26-Jul-2018.

\bibitem{Bojanowski2016EnrichingInformation}
Piotr Bojanowski, Edouard Grave, Armand Joulin, and Tomas Mikolov.
\newblock {Enriching word vectors with subword Information}.
\newblock {\em Transactions of the Association of Computational Linguistics},
  5(1):135--146, June 2017.

\bibitem{Grave2018LearningLanguages}
Edouard Grave, Piotr Bojanowski, Prakhar Gupta, Armand Joulin, and Tomas
  Mikolov.
\newblock {Learning word vectors for 157 languages}.
\newblock In {\em Proc. of the 11th Language Resources and Evaluation
  Conference}, pages 3483--3487. Proc. of the Eleventh International Conference
  on Language Resources and Evaluation (LREC-2018), Feb. 2018.

\bibitem{multilingual2016}
Igor Mozeti\v{c}, Miha Gr\v{c}ar, and Jasmina Smailovi\'{c}.
\newblock {Multilingual Twitter sentiment classification: The role of human
  annotators}.
\newblock {\em PloS one}, 11(5):1--26, May 2016.

\bibitem{CorpusArabic:2015}
NRC.
\newblock {Syrian tweets arabic sentiment analysis dataset}.
\newblock \url{http://saifmohammad.com/WebPages/ArabicSA.html}, 2017.
\newblock Accessed 17-Feb-2017.

\bibitem{Wilcoxon1945IndividualMethods}
Frank Wilcoxon.
\newblock {Individual comparisons by ranking methods}.
\newblock {\em Biometrics Bulletin}, 1(6):80, Dec. 1945.

\bibitem{Holm1979AProcedure}
Sture Holm.
\newblock {A simple sequentially rejective multiple test procedure}.
\newblock {\em Scandinavian Journal of Statistics}, 6:65--70, Dec. 1979.

\bibitem{Priban2018UWBTweets}
Priban Pavel, Tomas Hercig, and Ladislav Lenc.
\newblock {UWB at SemEval-2018 Task 1: Emotion intensity detection in tweets}.
\newblock In {\em Proc. of the 12th International Workshop on Semantic
  Evaluation}, pages 133--140. ACL, June 2018.

\bibitem{badaro-EtAl:2018EMA}
Gilbert Badaro, Obeida El~Jundi, Alaa Khaddaj, Alaa Maarouf, Raslan Kain, Hazem
  Hajj, and Wassim El-Hajj.
\newblock {EMA at SemEval-2018 Task 1: Emotion mining for Arabic}.
\newblock In {\em Proc. of The 12th International Workshop on Semantic
  Evaluation}, pages 236--244, Stroudsburg, PA, USA, June 2018. ACL.

\bibitem{rozental-fleischer:2018Amobee}
Alon Rozental and Daniel Fleischer.
\newblock {Amobee at SemEval-2018 Task 1: GRU neural network with a CNN
  attention mechanism for sentiment classification}.
\newblock In {\em Proc. of The 12th International Workshop on Semantic
  Evaluation}, pages 218--225. ACL, June 2018.

\bibitem{gee-wang:2018psyML}
Grace Gee and Eugene Wang.
\newblock {psyML at SemEval-2018 Task 1: Transfer learning for sentiment and
  emotion analysis}.
\newblock In {\em Proc. of The 12th International Workshop on Semantic
  Evaluation}, pages 369--376. ACL, June 2018.

\bibitem{karasalo-EtAl:2018FOIDSS}
Maja Karasalo, Mattias Nilsson, Magnus Rosell, and Ulrika Wickenberg~Bolin.
\newblock {FOI DSS at SemEval-2018 Task 1: Combining LSTM states, embeddings,
  and lexical features for affect analysis}.
\newblock In {\em Proc. of The 12th International Workshop on Semantic
  Evaluation}, pages 109--115. ACL, June 2018.

\bibitem{meisheri-dey:2018TCSResearch}
Hardik Meisheri and Lipika Dey.
\newblock {TCS research at SemEval-2018 Task 1: Learning robust representations
  using multi-attention architecture}.
\newblock In {\em Proc. of The 12th International Workshop on Semantic
  Evaluation}, pages 291--299. ACL, June 2018.

\bibitem{baziotis-EtAl:2018NTUA-SLP}
Christos Baziotis, Athanasiou Nikolaos, Alexandra Chronopoulou, Athanasia
  Kolovou, Georgios Paraskevopoulos, Nikolaos Ellinas, Shrikanth Narayanan, and
  Alexandros Potamianos.
\newblock {NTUA-SLP at SemEval-2018 Task 1: Predicting affective content in
  tweets with deep attentive RNNs and transfer learning}.
\newblock In {\em Proc. of The 12th International Workshop on Semantic
  Evaluation}, pages 245--255. ACL, June 2018.

\bibitem{davalfrerot-bouchekif-moreau:2018Epita}
Guillaume Daval-Frerot, Abdesselam Bouchekif, and Anatole Moreau.
\newblock {Epita at SemEval-2018 Task 1: Sentiment analysis using transfer
  learning approach}.
\newblock In {\em Proc. of The 12th International Workshop on Semantic
  Evaluation}, pages 151--155. ACL, June 2018.

\bibitem{Gonzalez2018ELiRF-UPVTweets}
Jos\'e-A Gonz\'{a}lez, Llu\'is-F Hurtado, and Ferran Pla.
\newblock {ELiRF-UPV at SemEval-2018 Tasks 1 and 3: Affect and irony detection
  in tweets}.
\newblock In {\em Proc. of the 12th International Workshop on Semantic
  Evaluation}, pages 565--569. ACL, June 2018.

\bibitem{Abdullah2018TeamUNCCLearning}
Malak Abdullah and Samira Shaikh.
\newblock {TeamUNCC at SemEval-2018 Task 1: Emotion detection in English and
  Arabic tweets using deep learning}.
\newblock In {\em Proc. of the 12th International Workshop on Semantic
  Evaluation}, pages 350--357. ACL, June 2018.

\bibitem{zhang-wang-zhang:2018YNU-HPCC}
You Zhang, Jin Wang, and Xuejie Zhang.
\newblock {YNU-HPCC at SemEval-2018 Task 1: BiLSTM with attention based
  sentiment analysis for affect in tweets}.
\newblock In {\em Proc. of The 12th International Workshop on Semantic
  Evaluation}, pages 273--278. ACL, June 2018.

\bibitem{kuijper-vanlenthe-vannoord:2018UG18}
Marloes Kuijper, Mike van Lenthe, and Rik van Noord.
\newblock {UG18 at SemEval-2018 Task 1: Generating additional training data for
  predicting emotion intensity in Spanish}.
\newblock In {\em Proc. of The 12th International Workshop on Semantic
  Evaluation}, pages 279--285. ACL, June 2018.

\bibitem{Pedregosa2011Scikit-learn:Python}
Fabian Pedregosa, Gaël Varoquaux, Alexandre Gramfort, Vincent Michel, Bertrand
  Thirion, Olivier Grisel, Mathieu Blondel, Peter Prettenhofer, Ron Weiss,
  Vincent Dubourg, Jake Vanderplas, Alexandre Passos, David Cournapeau,
  Matthieu Brucher, Matthieu Perrot, and Édouard Duchesnay.
\newblock {Scikit-learn: Machine learning in Python}.
\newblock {\em Journal of Machine Learning Research}, 12:2825--2830, Oct. 2011.

\end{thebibliography}
\end{document}